\documentclass{article}
\usepackage{ijcai20}

\usepackage{times}
\usepackage{soul}
\usepackage{url}
\usepackage[hidelinks]{hyperref}
\usepackage[utf8]{inputenc}
\usepackage[small]{caption}
\usepackage{graphicx}
\usepackage{amsmath}
\usepackage{amsthm}
\usepackage{booktabs}
\usepackage{algorithm}
\usepackage{algorithmic}
\usepackage{amssymb}

\usepackage{subfigure}
\usepackage{color}

\title{Soft Threshold Ternary Networks}

\author{
Weixiang Xu$^{1,2}$\and
Xiangyu He$^{1,2}$\and
Tianli Zhao$^{1,2}$\and
Qinghao Hu$^{1}$\and
Peisong Wang$^{1}$\And
Jian Cheng$^{1}$\\
\affiliations
$^1$NLPR, Institute of Automation, Chinese Academy of Sciences\\
$^2$School of Artificial Intelligence, University of Chinese Academy of Sciences
\\
\emails
\{xuweixiang2018,hexiangyu2017,zhaotianli2019\}@ia.ac.cn,
\{qinghao.hu,peisong.wang,jcheng\}@nlpr.ia.ac.cn,
}

\begin{document}

\maketitle

\begin{abstract}
  Large neural networks are difficult to deploy on mobile devices because of intensive computation and storage. To alleviate it, we study ternarization, a balance between efficiency and accuracy that quantizes both weights and activations into ternary values. In previous ternarized neural networks, a hard threshold $\Delta$ is introduced to determine quantization intervals. Although the selection of $\Delta$ greatly affects the training results, previous works estimate $\Delta$ via an approximation or treat it as a hyper-parameter, which is suboptimal. In this paper, we present the Soft Threshold Ternary Networks (STTN), which enables the model to automatically determine quantization intervals instead of depending on a hard threshold. Concretely, we replace the original ternary kernel with the addition of two binary kernels at training time, where ternary values are determined by the combination of two corresponding binary values. At inference time, we add up the two binary kernels to obtain a single ternary kernel. Our method dramatically outperforms current state-of-the-arts, lowering the performance gap between full-precision networks and extreme low bit networks. Experiments on ImageNet with ResNet-18 (Top-1 66.2$\%$) achieves new state-of-the-art.

  \textcolor{red}{Update:} In this version, we further fine-tune the experimental hyperparameters and training procedure. The latest STTN shows that ResNet-18 with ternary weights and ternary activations achieves up to 68.2$\%$ Top-1 accuracy on ImageNet. Code is available at:
  \url{github.com/WeixiangXu/STTN}.
\end{abstract}

\section{Introduction}

Recently, deep convolution neural networks have made significant improvements in lots of fields, including but not limited to computer vision~\cite{krizhevsky2012imagenet,lin2017focal}, speech recognition, and natural language processing. Attracted by the excellent performance of CNN models, many people try to deploy CNNs to real-world applications. However, the superior performance of DNN usually requires powerful hardware with abundant computing and memory resources, for example, high-end graphics processing units (GPUs).

\begin{figure}[t]
\flushright
\includegraphics[width=3in]{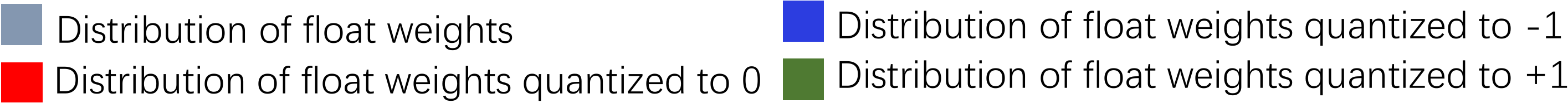}
\centering
\includegraphics[height=2.36in]{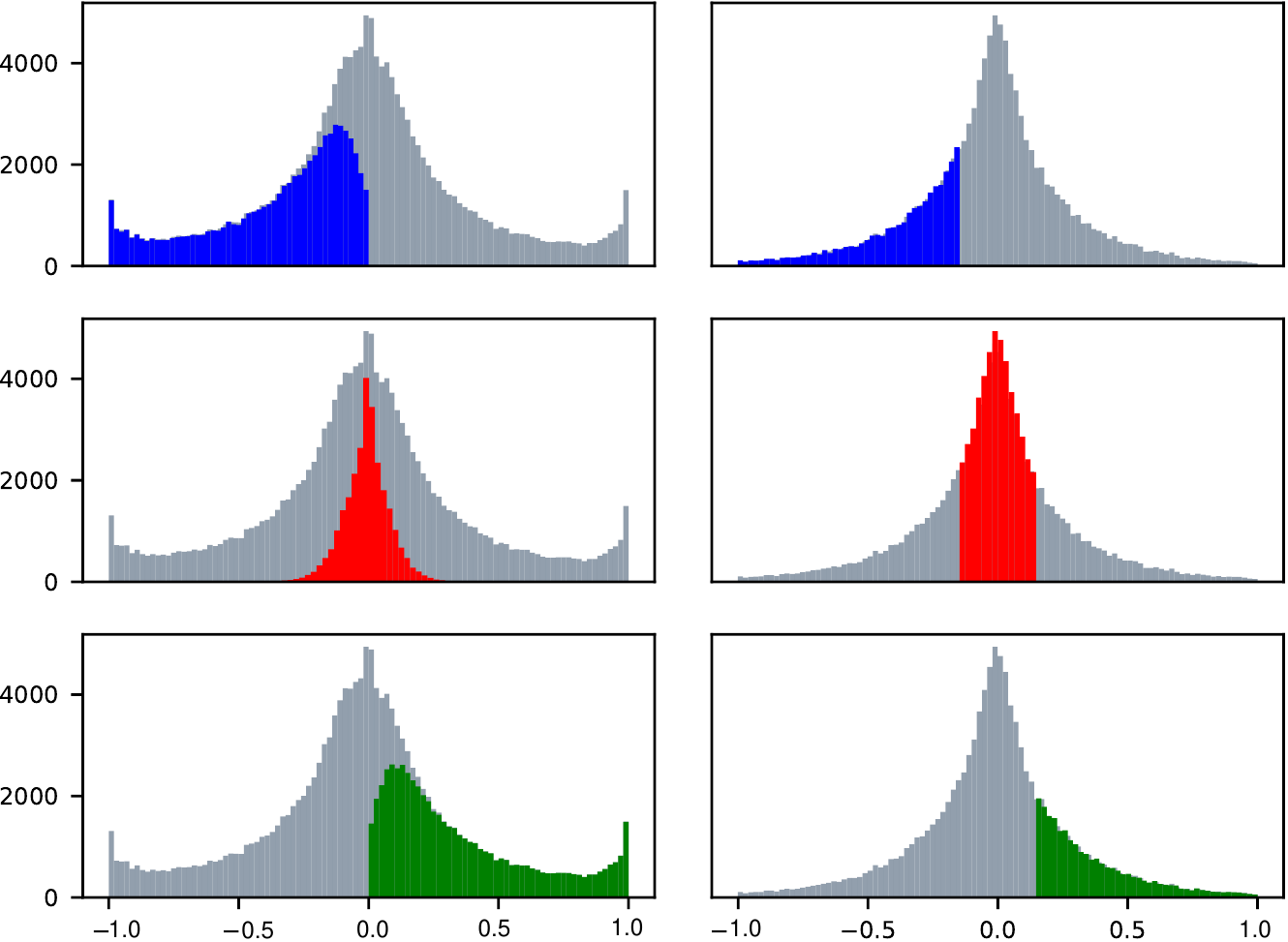}
\centering
\includegraphics[width=3.15in]{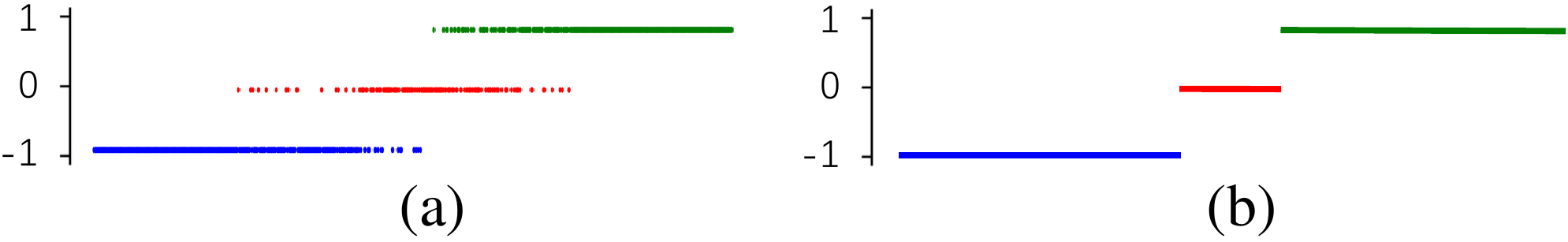}
\caption{Comparison between \textit{soft} and \textit{hard} threshold. The weight distributions are drawn from ResNet-18 on ImageNet. We take Layer1.0.conv1 as an example. (a) Our soft threshold ternarization. (b) Hard threshold ternarization which splits intervals with $\Delta$. The blue/red/green distribution comes from floating point weights quantized to -1/0/1, respectively. The blue/red/green lines below distributions denote positions of floating point weights that will be quantized to -1/0/1 when sorting weights from small to large.}
\label{fig:soft_threshold}
\end{figure}

Deploying deep neural networks to resource limited devices is still a challenging problem due to the requirements of abundant computing and memory resources.  A variety of methods have been proposed to reduce the parameter size and accelerate the inference phase, such as compact model architecture design (in a handcrafted way or automatic search way)~\cite{howard2017mobilenets,zoph2016neural}, network pruning~\cite{han2015deep,he2019filter}, knowledge distilling~\cite{hinton2015distilling}, low-rank approximation~\cite{jaderberg2014speeding,jaderberg2014speeding}, etc. Besides, recent works~\cite{courbariaux2014low,gupta2015deep} show that full-precision weights and activations are not necessary for networks to achieve high performances. This discovery indicates that both weights and activations in neural networks can be quantized to
low-bit formats. In this way, both storage and computation resources can be saved. 

The extreme cases of network quantization are binary and ternary quantization ($\{-1,1\}$ and $\{-1,0,1\}$). In computationally heavy convolutions, multiply-accumulate consumes most of the operation time. Through binarization or ternarization, multiply-accumulate can be replaced by cheap bitwise $xnor$ and $popcount$ operations~\cite{courbariaux2016binarized}. 
Although binary networks can achieve high compression rate and computing efficiency, they inevitably suffer from a long training procedure and noticeable accuracy drops owing to poor representation ability.

In order to make a balance between efficiency and accuracy, ternary CNNs convert both weights and activations into ternary values. Theoretically, ternary CNNs have stronger representation ability than binary CNNs, and should have a better performance. Although many works~\cite{li2016ternary,zhu2016trained,wang2018two,wan2018tbn} provide tremendous efforts on ternary neural networks, we argue that there are two main issues in existing works:
1) All previous ternary networks use a \textit{hard} $\Delta$ to divide floating point weights into three intervals (see Eq.\ref{eq:wt} for more details), which introduces an additional constraint into ternary networks.
2) Although the theoretical optimal $\Delta$ can be calculated through optimization, calculating it at each training iteration is too time-consuming. Therefore, previous methods either use a hyper-parameter to estimate $\Delta$ or only calculate the exact $\Delta$ at the first training iteration then keep it for later iterations, which further limits ternary network accuracy. A brief summary: they introduce an additional constraint based on $\Delta$. Further more, a suboptimal $\Delta$ is used instead of the exact one.

In this paper, we propose Soft Threshold Ternary Networks (STTN), an innovative method to train ternary networks. STTN avoids the \textit{hard} threshold and enables the model to automatically determine which weights to be $-1/0/1$ in a \textit{soft} manner.
We convert training ternary networks into training equivalent binary networks (more details in section 3.2). STTN enjoys several benefits: 1) The constraint based on $\Delta$ is removed. 2) STTN is free of calculation of $\Delta$. 3) As shown in Figure~\ref{fig:soft_threshold} (a), weights are ternarized in a soft manner, which is why we name it \textit{soft} threshold.

Our contributions are summarized as follows.
\begin{itemize}
\item We divide previous ternary networks into two catalogues: Optimization-based method and Learning-based method. Analysis is given about issues in the existing ternary networks that prevent them from reaching high performance.
\item We propose STTN, an innovative way to train ternary networks by decomposing one ternary kernel into two binary kernels during training. No extra parameters or computations are needed during inference. Quantizing to -1/0/1 is determined by two binary variables rather than decided by a hard threshold.
\item We show that our proposed ternary training method provides competitive results on the image classification, i.e. CIFAR-10, CIFAR-100 and ImageNet datasets. Qualitative and quantitative experiments show that it outperforms previous ternary works.
\end{itemize}

\section{Related Work}
Low-bit quantization of deep neural networks has recently received increasing interest in deep learning communities. 
By utilizing fixed-point weights and feature representations, not only the model size can be dramatically reduced, but also inference time can be saved.

The extreme cases of network quantization are binary neural networks (BNN) and ternary neural networks (TNN). The runtime speed of BNN/ TNN can be comparable with GPU by replacing expensive floating-point matrix multiplication with high computation efficiency POPCOUNT-XNOR operations~\cite{rastegari2016xnor}. And our study of ternary networks touches upon the following areas of research.

\subsection{Binary Networks}

BNN~\cite{courbariaux2016binarized} constrains both the weights and activations to either +1 or -1, which produces reasonable results on small datasets, such as MNIST and CIFAR-10. However, there is a significant accuracy drop on large scale classification datasets, such as ImageNet. Some improvements based on BNN have been investigated. For example,~\cite{darabi2018bnn+} introduces a regularization function that encourages training weights around binary values.~\cite{tang2017train} proposes to use a low initial learning rate. Nevertheless, there is still a non-negligible accuracy drop. To improve the quality of the binary feature representations, XNOR-Net~\cite{rastegari2016xnor} introduces scale factors for both weights and activations during binarization process. DoReFa-Net~\cite{zhou2016dorefa} further improves XNOR-Net by approximating the activations with more bits. Since ABC-Net~\cite{lin2017towards}, several works propose to decompose a single convolution layer into K binary convolution operations~\cite{gu2019bonn,zhu2019binary}. Although higher accuracy can be achieved, K$\times$ extra parameters and computations are needed for both training and inference time, which defeats the original purpose of binary networks. Different from those K$\times$ binary methods, we decompose a ternary convolution layer into two parallel binary convolution layers only during training. Once training finishes, we only store the ternary model and use the ternary model for inference.

\subsection{Ternary Networks}

To make a balance between efficiency and accuracy, ternary CNNs convert both weights and activations into ternary values. Cheap bitwise operations can still be used to save inference time in this case. TWN~\cite{li2016ternary} ternarize neural networks weights to be $\{+1, 0, -1\}$ by minimizing Euclidian distance between full precision weights and the ternary weights along with a scaling factor. TTQ~\cite{zhu2016trained} set scaling factors as learnable parameters to achieve higher performance. TSQ~\cite{wang2018two} train a full precision network at first, then initialize ternary weights with pretrained parameters through solving an optimization problem. RTN~\cite{li2019rtn} proposes to reparameter activation and weights by introducing extra learnable parameters. To tackle the challenge that the complexity of ternary inner product is the same as the standard 2-bit counterparts,
FATNN~\cite{chen2021fatnn} proposes a new ternary quantization pipeline, which reduces the complexity of TNNs by 2$\times$.
We will divide those methods into two catalogues and analyse the unresolved/overlooked issues in previous works.
We will review previous ternary works in detail in section 3.1.

\section{Methodology}

We first revisit the formulation of previous ternary networks. We divide them into two catalogues and show their common issues in calculating the appropriate $\Delta$. 
We then present our novel Soft Threshold Ternary Networks in detail, including scaling coefficient constraint and backward approximation. We show that our method can avoid the previous issue in a simple but effective way.

\subsection{Review of Previous Ternary Networks}
\subsubsection{Problem Formulation}
Full precision convolution or inner-product can be formed as: $Y=\sigma(W^TX)$. Here $\sigma(\cdot)$ represents the nonlinear activation function, such as ReLU. $W\in\mathbb{R}^{n\times{chw}}$ and $X\in\mathbb{R}^{chw\times{HW}}$ are float point weights and inputs respectively, where $(n, c, h, w)$ are filter number, input channels, kernel height and kernel width, and $(H,W)$ are height and width of output feature maps. Ternary networks convert both weights and inputs into ternary values: $T\in\{+1, 0, -1\}^{n\times{chw}}$ and $X^t\in\{+1, 0, -1\}^{chw\times{HW}}$. 

As for weights ternarization, 
$\alpha$ is used to estimate the floating weight $W$ alone with ternary weight $T$.
\begin{equation}
    W\approx\alpha T
\end{equation}

Previous works formulate ternarization as a weight approximation optimization problem:
\begin{equation}
    \begin{cases}
    \alpha^*,T^*=   & \mathop{\arg\min}\limits_{\alpha,T}
                     J(\alpha, T) = \Vert W-\alpha T\Vert_2^2   \\
    \qquad s.t.   & \alpha\ge0, T_i \in \{-1, 0, 1\}, i = 1,2,...,n. \\
    \end{cases}
\label{eq:opt1}
\end{equation}%

A \textit{hard} threshold $\Delta$ is then introduced by previous works to divide quantization intervals, which sets an additional constraint on ternary networks: 
\begin{equation}
    T_i=
    \begin{cases}
    +1,& \text{if $W_i>\Delta$}\\
    0,& \text{if $\left|W_i\right|\le \Delta$}\\
    -1,& \text{if $W_i<-\Delta$}
    \end{cases}
\label{eq:wt}
\end{equation}%

As illustrated in Figure \ref{fig:soft_threshold}(b), $\Delta$ plays a role as hard threshold here, splitting the floating point weights into $\{-1,0,1\}$. We will show that the above constraint can be avoided in our method. According to methods of calculating $\alpha$ and $\Delta$, we divide previous ternary networks into 1) optimization-based ternary networks, and 2) learning-based ternary networks.
\subsubsection{Optimization-based Ternary Networks}
TWN~\cite{li2016ternary} and TSQ~\cite{wang2018two} formulate the ternarization as Eq.(\ref{eq:opt1}).

In TWN, $T$ is obtained according to threshold-based constraint Eq.(\ref{eq:wt}). They give exact optimal $\alpha^*$ and $\Delta^*$ as follows:
\begin{equation}
\resizebox{.91\linewidth}{!}{$
    \alpha^*_{\Delta}= \frac{1}{|I_\Delta|}\sum\limits_{i\in I_\Delta}|W_i|;~~
    \Delta^* = \mathop{\arg\max}\limits_{\Delta>0} \frac{1}{|I_\Delta|}(\sum\limits_{i\in I_\Delta} |W_i|)^2
$}
\label{eq:solution1}
\end{equation}%
where $I_\Delta=\{i\big||W_i|>\Delta\}$ and $|I_\Delta|$ denotes the number of elements in $I_\Delta$.
Note that $\Delta^*$ in Eq.(\ref{eq:solution1}) has no straightforward solution. It can be time and computing consuming if using discrete optimization to solve it. In TWN, they adopt an estimation $\Delta '\approx 0.7\cdot E(|W|) $ to approximate the optimal $\Delta^*$. Obviously there is a gap between $\Delta^*$ and $\Delta '$. What's more, the optimal scaling factor $\alpha^*$ in Eq.(\ref{eq:solution1}) will also be affected because of dependence on $\Delta$.

In TSQ, they try to obtain $T^*$ by directly solving Eq.(\ref{eq:opt1}). They give exact optimal $\alpha^*$ and $T^*$ as follows\footnote{$W^T$ denotes transpose of matrix $W$. $T$ denotes ternary weights.}:
\begin{equation}
    \alpha^*= \frac{W^T T}{T^T T};~~~
    T^* = \arg\max\limits_{T}\frac{(W^TT)^2}{T^T T}
\label{eq:solution2}
\end{equation}

Note that there is still no straightforward solution for $T^*$ in Eq.(\ref{eq:solution2}). TSQ proposes OTWA(Optimal Ternary Weights Approximation) to obtain $T^*$. However, sorting for $|W_i|$ is needed in their algorithm. Time complexity is higher than $O(N\log(N))$, where $N$ is the number of elements in kernel weights. To avoid time consuming sorting, they only calculate $\Delta$ at the first training iteration and then keep it for later iterations, which is suboptimal.

\subsubsection{Learning-based Ternary Networks}

TTQ~\cite{zhu2016trained} tries another way to obtain scaling factor $\alpha$ and threshold $\Delta$. They set $\alpha$ as a learnable parameter and set $\Delta$ as a hyper-parameter. Based on TTQ, RTN~\cite{li2019rtn} introduces extra float point scaling and offset parameters to reparameterize ternary weights.

Because the search space of hyper-parameter $\Delta$ is too large, both of TTQ and RTN set $\Delta$ the same across all layers, e.g. $\Delta=0.5$ in RTN. We argue that ternary methods with hyper-parameter are not reasonable as following:
\begin{figure}[t]
\includegraphics[width=3.33in]{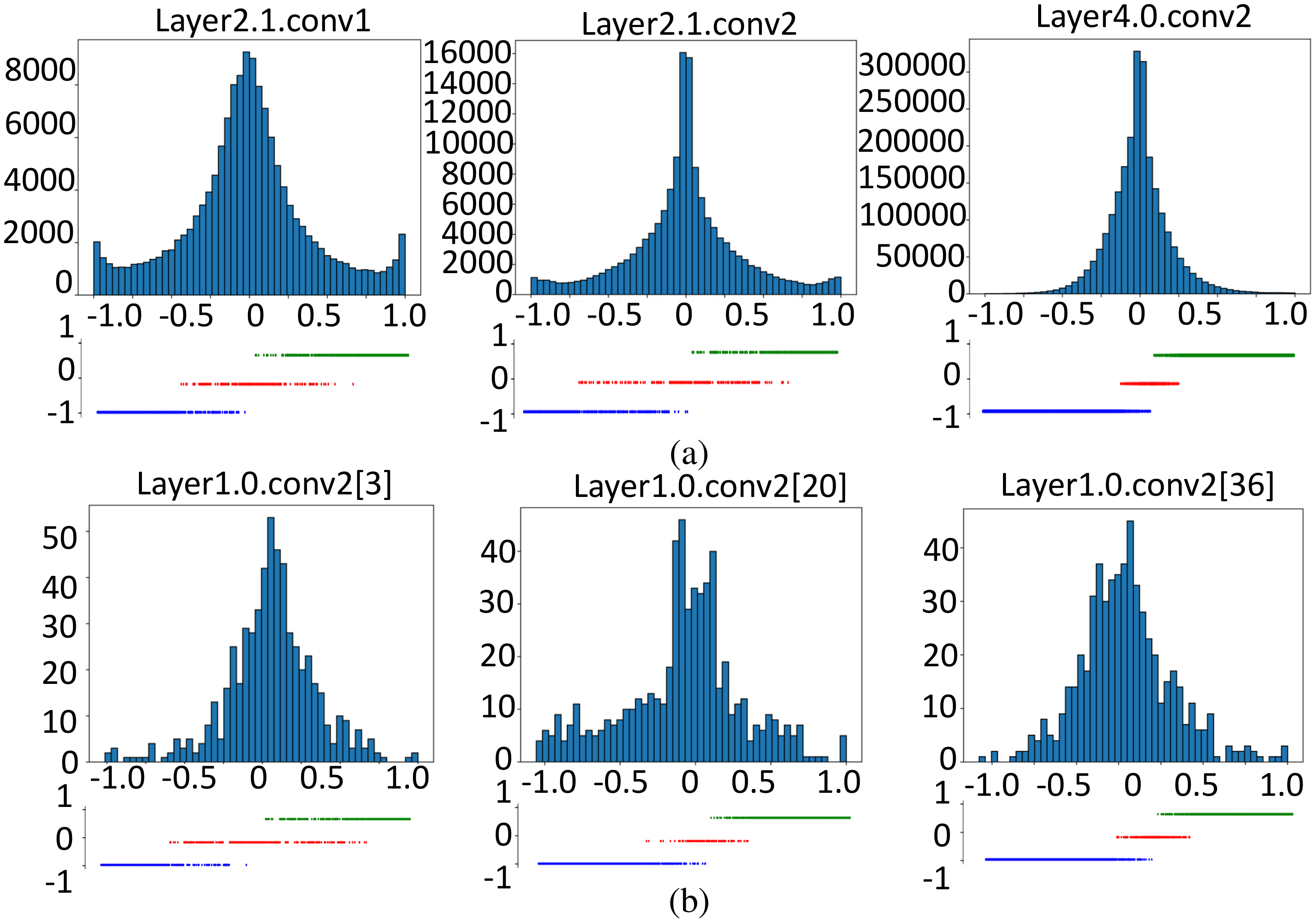}
\caption{Weight distribution of ResNet-18 on ImageNet. (a) Distributions from different layers. (b) Distributions from different kernels in the same layer. "Layer1.0.conv2[36]" denotes the 37th kernel in Layer1.0.conv2. The blue/red/green lines below each distribution denote positions of floating point weights that will be quantized to -1/0/1, which are obtained through our STTN.}
\label{fig:weight_distribution}
\end{figure}

In Figure~\ref{fig:weight_distribution}, we visualize the weight distribution of ResNet-18 on ImageNet from different layers and from different kernels in the same layer, respectively. From the figure, we can observe that the distributions are different from layer to layer. Even for kernels in the same layer, their weight distributions can be diverse. Therefore, 1) it is not reasonable for previous works to set a fixed threshold for all kernels (e.g. $\Delta$ = 0.5 in TTQ and RTN). 2) What's more, their ternarization also use hard threshold: once $\Delta$ is set, their quantized intervals are determined according to Eq.(\ref{eq:wt}).

So can we remove the hard $\Delta$ constraint Eq.(\ref{eq:wt}), solving ternary networks in another way?
\subsection{Soft Threshold Ternary Networks}
Due to the issues analyzed above, we propose our novel STTN. Our motivation is to enable the model to automatically determine which weights to be -1/0/1, avoiding the hard threshold $\Delta$.

We introduce our methods via convolution layers. The inner-product layer has a similar form. Concretely, at \textit{training} time, we replace the ternary convolution filter $T$ with two parallel binary convolution filters $B_1$ and $B_2$. They are both binary-valued and have the same shape with ternary filter: $n\times chw$. Due to the additivity of convolutions with the same kernel sizes, a new kernel can be obtained by:
\begin{equation}
    T = B_1 + B_2
\end{equation}

\begin{figure}[t]
\includegraphics[height=1.5in]{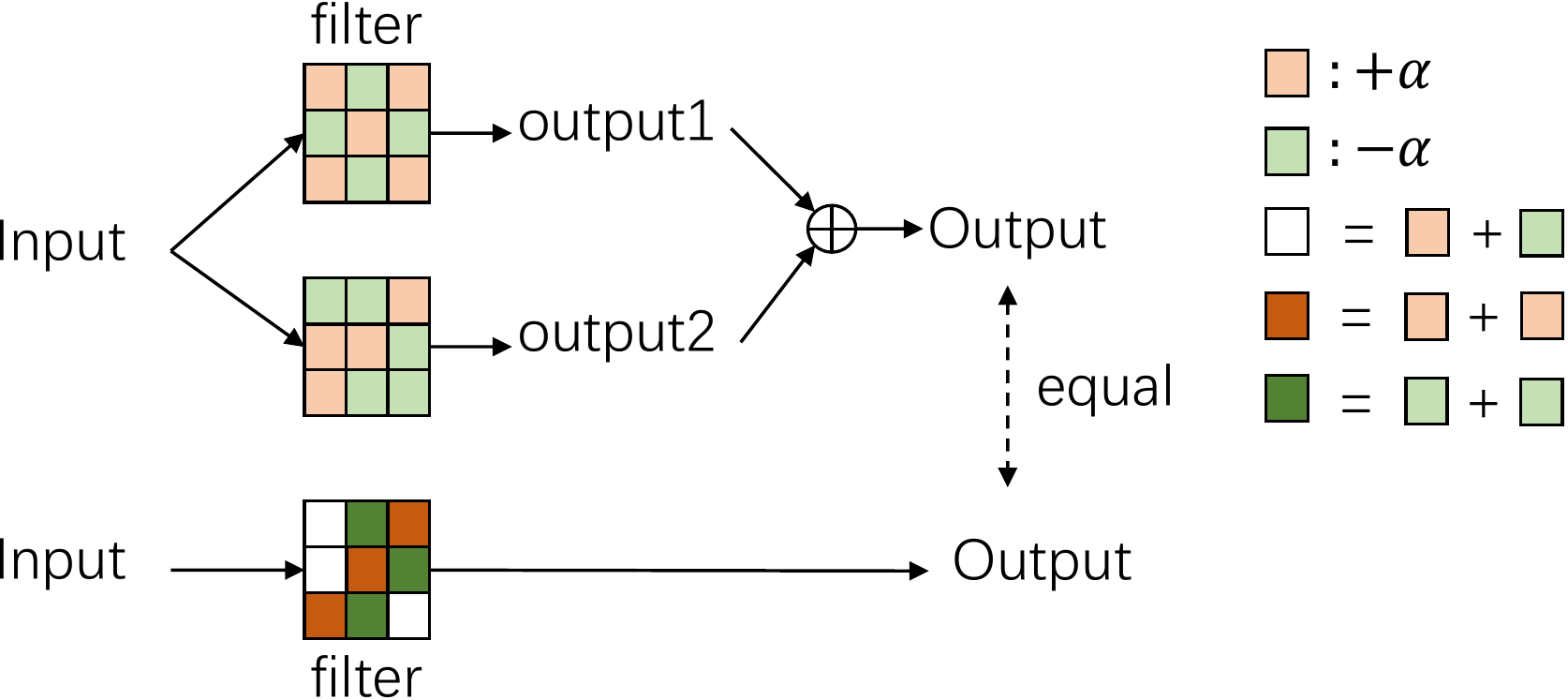}
\caption{We illustrate our method with a 2D convolution for simplicity. The top of the figure are two training-time binary convolution kernels. Their weights are binary-valued:$\{+\alpha, -\alpha\}$. The bottom of the figure is an inference-time ternary convolution kernel who takes the same input. The inference-time kernel can be obtained by easily adding the corresponding two binary kernels. The weights are ternary-valued: $\{+2\alpha, 0, -2\alpha\}$. The additivity of convolutions tells us: the inference-time model can produce the same outputs as the training-time.}
\label{fig:add}
\end{figure}

A key requirement for $T$ to have ternary values is that those two binary filters should have the same scaling factors: $\alpha$. With $\alpha B_1,\alpha B_2 \in \{+\alpha,-\alpha \}^{n\times chw}$, the sum of $\alpha B_1$ and $\alpha B_2$ is ternary-valued, i.e. $\alpha T \in \{+2\alpha, 0, -2\alpha\}^{n\times{chw}}$. Note that we only decompose the filters at training time. After the training, the two trained parallel binary filters are added up to obtain the ternary filter. Thus there is no extra computation when deploying trained models to devices. An example is illustrated in Figure \ref{fig:add}.

Zeroes are introduced in ternary filters (white squares in Figure \ref{fig:add} at positions where two parallel filters have opposite values). And -1/1 is obtained at positions where two parallel filters have the same signs. In this way, ternary values are determined by the combination of two corresponding binary values.

It is obvious that the outputs of training-time model are equal to the outputs of inference-time model:
\begin{equation}
    Y=\sigma ( \alpha B_1^T X + \alpha B_2^T X)
    =\sigma(\alpha T^T X)
\end{equation}

In our proposed STTN, we convert training ternary networks into training equivalent binary networks. We abandon the constraint in Eq.(\ref{eq:wt}). Not only $\Delta$ can be avoided, but also quantization intervals can be divided in a soft manner.

\subsubsection{Scaling Coefficient}
As mentioned above, a key requirement is that the corresponding two parallel binary filters should have the same scaling factors to guarantee the sum of them are ternary. Taking the two parallel binary filters into consideration, we obtain the appropriate $\{\alpha_1, \alpha_2\}$ by minimizing the following weight approximation problem:
\begin{equation}
\begin{aligned}
    J(\alpha_1,\alpha_2,B_1,B_2) &= \Vert W_1- \alpha_1 B_1 \Vert^2_2
    + \Vert W_2- \alpha_2 B_2 \Vert^2_2         \\
    &s.t. ~~\alpha_1 = \alpha_2; ~\alpha_1, \alpha_2 \ge 0
\label{eq:cal_alpha}
\end{aligned}
\end{equation}

Here $B_1$ and $B_2$ are the two parallel binary filters. And $W_1$ and $W_2$ are the corresponding float point filters. With constraint $\alpha_1 = \alpha_2$, we use $\alpha$ to denote them. Through expanding Eq.(\ref{eq:cal_alpha}), we have
\begin{equation}
\resizebox{.99\linewidth}{!}{$
\begin{aligned}
    J(\alpha_1,\alpha_2,B_1,B_2) =
    \alpha^2(B_1^T B_1 +B_2^T B_2)
    -2\alpha(B_1^T W_1 + B_2^T W_2) + C
\label{eq:alpha_b}
\end{aligned}
$}
\end{equation}

$C=W_1^TW_1 + W_2^TW_2$ is a constant because $W_1$ and $W_2$ are known variables. In order to get $\alpha^*$, the optimal $B_1^*$ and $B_2^*$ should be determined. Since $B_1,B2\in \{+1,-1\}^{n\times chw}$, $B_1^T B_1+B_2^T B_2 = 2nchw$ is also a constant. From Eq.(\ref{eq:alpha_b}), $B_1^*$ and $B_2^*$ can be achieved by maximizing $B_1^T W_1 + B_2^T W_2$ with constraint condition that $B_1,B_2 \in \{+1,-1\}^{n\times chw}$. Obviously the optimal solution can be obtained when binary kernel has the same sign with the corresponding float point kernel at the same positions, i.e. $B_1^*=sign(W_1)$, $B_2^*=sign(W_2)$.

Based on the optimal $B_1^*$ and $B_2^*$, $\alpha^*$ can be easily calculated as:
\begin{equation}
\resizebox{.89\linewidth}{!}{$
\begin{aligned}
    \alpha^* = \frac{B_1^T W_1 +B_2^T W_2}{B_1^T B_1 + B_2^T B_2}
    =\frac{1}{2N}(\sum_{i=1}^N |W_{1i}| + \sum_{i=1}^N |W_{2i}|)
\label{eq:my_alpha}
\end{aligned}
$}
\end{equation}
where $N=nchw$, is the number of elements in each weight. $W_{1i}$ and $W_{2i}$ are elements of $W_{1}$ and $W_{2}$, respectively.

\subsubsection{Backward Approximation}

Since we decompose one ternary filter into two parallel binary filters at training time, binary weights approximation is needed in both forward and backward processes. During the forward propagation, the two related weights can be binarized through sign function along with the same scaling factor calculated by Eq.(\ref{eq:my_alpha}). However, during the backward propagation, the gradient of sign function is almost everywhere zero. Assume $\ell$ as the loss function and $\widetilde{W}=\alpha B=\alpha \cdot sign(W)$ as the approximated weights, XNOR-Net~\cite{rastegari2016xnor} alleviates this problem through Straight-Through Estimator (STE):
\begin{equation}
\begin{aligned}
    \frac{\partial\ell}{\partial W_i} = \frac{\partial \ell}{\partial \widetilde{W_i}}\frac{\partial \widetilde{W_i}}{\partial W_i}    = \frac{\partial \ell}{\partial \widetilde{W_i}}\Big(\frac{1}{N}+ \alpha \frac{\partial sign(W_i)}{\partial W_i}\Big)
\label{eq:xnor_derivate_w}
\end{aligned}
\end{equation}

Here, $sign(W_i)$ is approximated with $W_i\boldsymbol{1}_{|W_i| \le 1}$. $N$ is the number of elements in each weight.

However, note that an important requirement in our STTN is that the related two parallel binary filters should have the same scaling factors. The exact approximated weights should be:
\begin{equation}
\widetilde{W}=\alpha B= \frac{1}{2N}(\sum_{i=1}^N |W_{1i}| + \sum_{i=1}^N |W_{2i}|) \cdot sign(W)
\end{equation}
Because Eq.(\ref{eq:my_alpha}) indicates that $\alpha$ is dependent on both $W_1$ and $W_2$. When calculating the derivatives of $W_{1i}$, the effect of other kernels $W_{1j}$ and $W_{2j}$ should be considered. But Eq.(\ref{eq:xnor_derivate_w}) ignores the effect of $W_{1j}$ and $W_{2j}$, which is not suitable for our backward approximation.

Taking above analysis into consideration, we propose to calculate derivatives of $W$ in a more precise way:
\begin{equation}
\resizebox{.98\linewidth}{!}{$
\begin{aligned}
    \frac{\partial \ell}{\partial W_{1i}}
    &=\sum_{k=1}^{2} \sum_{j=1}^{N}
    \frac{\partial \ell}{\partial \widetilde{W}_{kj}} \frac{\partial \widetilde{W}_{kj}}{\partial W_{1i}}  \\
    &=\sum_{k=1}^{2} \sum_{j=1}^{N} \frac{\partial \ell}{\partial \widetilde{W}_{kj}} \Big[ \frac{1}{2N} \frac{\partial \vert W_{1i}\vert}{\partial W_{1i}} sign(W_{kj}) + \alpha \frac{\partial sign(W_{1i})}{\partial W_{1i}} \Big]
    \\
    &=\frac{1}{2N}sign(W_{1i}) \sum_{k=1}^{2} \sum_{j=1}^{N} \Big[ \frac{\partial \ell}{\partial \widetilde{W}_{kj}} sign(W_{kj})\Big]
    \\
    &{}+\alpha \frac{\partial sign(W_{1i})}{\partial W_{1i}} \sum_{k=1}^{2} \sum_{j=1}^{N} \frac{\partial \ell}{\partial \widetilde{W}_{kj}}
\end{aligned}
$}
\label{eq:gradient_of_W}
\end{equation}

Here $W_k=[W_{k1}, W_{k2},...,W_{kN}]$ (~$k\in\{1,2\}$) are the two parallel kernels respectively. 
$\partial \ell / \partial W_{2i}$ can be calculated in the same way as Eq.({\ref{eq:gradient_of_W}).

\subsubsection{Activation}
In this paper, we also convert activations into ternary values. We use the same ternarization function as RTN \cite{li2019rtn}. Given floating point activation $X$, the ternary activation is calculated by the following equation. The difference between ours and RTN is that we do not introduce extra parameters or calculations.
\begin{equation}
    X^t_i = Ternarize(X_i)=
    \begin{cases}
    sign(X_i),& \text{if $|X_i|>0.5$}   \\
    0,& \text{otherwise}                \\
    \end{cases}
\end{equation}


During the backward process, as previous binary/ternary works~\cite{courbariaux2016binarized,rastegari2016xnor,liu2018bi}, gradients propagated through ternarization function are estimated by Straight-Through Estimator (STE).





\section{Experiments}

In this section, we evaluate the proposed STTN in terms of qualitative and quantitative studies. Our experiments are conducted on three popular image classification datasets: CIFAR-10, CIFAR-100 and ImageNet (ILSVRC12). We test on several representative CNNs including: AlexNet, VGG-Net, and ResNet.

\subsection{Implementation Details}
We adopt the standard data augmentation scheme. In all CIFAR experiments, we pad 2 pixels in each side of images and randomly crop 32$\times$32 size from padded images during training. As for ImageNet experiments, we first proportionally resize images to 256$\times N$ ($N \times 256$) with the short edge to 256. Then we randomly sub-crop them to 224$\times$224 patches with mean subtraction and randomly flipping. No other data augmentation tricks are used during training.

Following RTN~\cite{li2019rtn}, we modify the block structure as BatchNorm $\rightarrow$ Ternarization $\rightarrow$ TernaryConv $\rightarrow$ Activation. Following XNOR-Net~\cite{rastegari2016xnor}, we place a dropout layer with $p = 0.5$ before the last layer for AlexNet. For VGG-Net, we use the same architecture VGG-7 as TWN~\cite{li2016ternary} and TBN~\cite{wan2018tbn}. We do not quantize the first and the last layer as previous binary/ternary works. We replace all $1\times1$ downsampling layers with max-pooling in ResNet.

We use Adam with default settings in all our experiments. The batch size for ImageNet is 256. We set weight decay as $1e^{-6}$ and momentum as 0.9. All networks on ImageNet are trained for 110 epochs. The initial learning rate is 0.005, and we use cosine learning rate decay policy. All our models are trained from scratch.
\subsection{Weight Approximation Evaluation}

\begin{table}
\centering
\begin{tabular}{lllr}
\toprule
Method                        &TWN     &TTQ     &Ours    \\
\midrule
$\Vert W-\alpha T\Vert_2^2$   &537.76  &439.46  &\textbf{379.25}\\
\bottomrule
\end{tabular}
\caption{$L2$ distance between approximated ternary weights and float point weights. (We add up distances of all convolution layers together.)}
\label{tab:weight_appro}
\end{table}

In this section, we explore the effect of the proposed STTN from qualitative view.

Previous works quantize weights into $\{-1,0,1\}$ by setting a hard threshold $\Delta$. Different from them, the proposed STTN 
generates soft threshold, quantizing weights more flexibly.
We illustrate the impact of threshold calculation on the performance of TNN based on ResNet-18. We first calculate the distance between trained floating weights $W$ and trained ternary weights $T$. $L2$ norm is used as the criterion for measurement as Eq.(\ref{eq:opt1}). We compare our method with TWN
and TTQ
. The results are shown in Table~\ref{tab:weight_appro}.

We can see that STTN obtains the smallest gap between trained floating weights and trained ternary weights, which indicates that our methods can realize a good weight approximation. Intuitively, the smaller the approximated weight error we get, the higher precision the model can obtain. The results show the effect of STTN qualitatively and further quantitative analyses are given in section 4.3.

We also analyze the reason why our STTN can obtain a smaller approximation error than previous ternary method. An essential effect of STTN is to quantize weights into $\{+1,0,-1\}$ in a soft manner, just as shown in Figure \ref{fig:soft_threshold}(a) and Figure \ref{fig:weight_distribution}. The effect comes from that STTN throws the hard threshold away. That is, Eq.(\ref{eq:wt}) is no more a constraint to the weight approximation optimization problem. By comparing Figure\ref{fig:soft_threshold}(a) with (b): this provides more flexible ternarization intervals.

Besides, we find that STTN adjust the weight sparsity regularly. Figure~\ref{fig:sparsity} shows the weight sparsity rates (the percentage of zeros) of different layers in our STTN on ResNet-18. From the figure we can see that the sparsity rates gradually decrease from the first layer to the last layer. This probably because high-level semantics need dense kernels to encode.
\begin{figure}[t]
\centering
\includegraphics[width=3.2in]{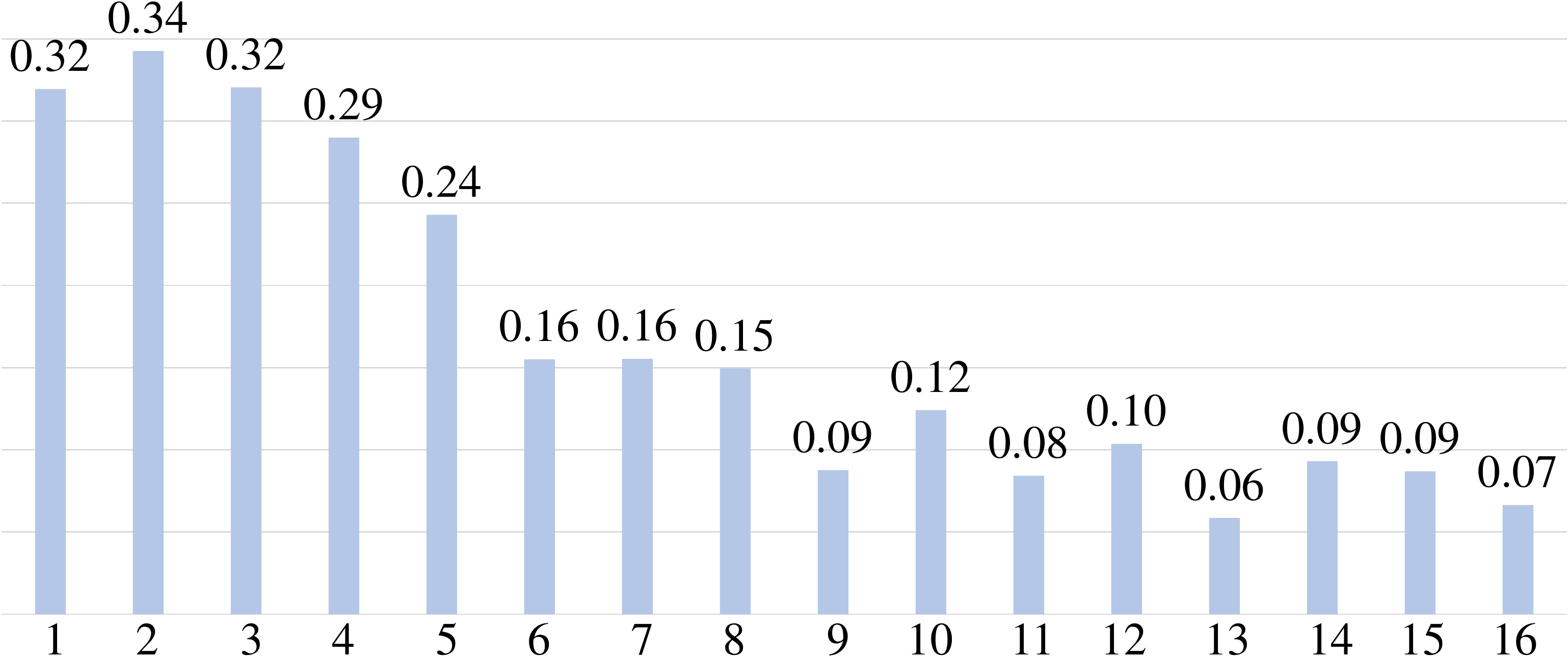}
\caption{Weight sparsity rate of different layer in our STTN on ImageNet with ResNet-18. Here we illustrate it with 16 convolution layers in building blocks of ResNet.}
\label{fig:sparsity}
\end{figure}
\subsection{Network Ternarization Results}
In this section, we evaluate the STTN from quantitative view by comparing with the state-of-the-art low-bit networks on various architectures. We ternarize both weights and activations. Experiments on only quantizing weights are also given.
\subsubsection{Results on CIFAR-10}
We first conduct experiments on CIFAR-10 dataset. We use the same network architecture as TWN, denoted as VGG-7. Compared with the architecture VGG-9 adopted in BNN and XNOR, the last two fully connection layers are removed. Table~\ref{tab:vgg7_cifar10} shows the STTN results. Note that for VGG-7, STTN with ternary weights and activations can even obtain better performance than the full-precision model.
\begin{table}
\centering
\begin{tabular}{lrr}
\toprule
Bit-width  &Method                      &Error($\%$)  \\
\midrule
32+32      &Floating point~\cite{li2016ternary}  &7.12       \\
\midrule
1+32$^\dag$       &BWN~\cite{courbariaux2015binaryconnect} &8.27  \\
1+1$^\dag$        &BNN~\cite{courbariaux2016binarized}     &10.15 \\
1+1$^\dag$        &XNOR~\cite{rastegari2016xnor}           &9.98  \\
2+32              &TWN~\cite{li2016ternary}                &7.44  \\
1+2               &TBN~\cite{wan2018tbn}                   &9.15  \\
\midrule
2+2               &Ours                                    &\textbf{7.07} \\
\bottomrule
\end{tabular}
\caption{The error rates on CIFAR-10 with VGG-7. The number before and after "+" in the first column denotes the weight and activation bit-width respectively. $\dag$ denotes the architecture is VGG-9, which adds two more FC layers at last.}
\label{tab:vgg7_cifar10}
\end{table}

\subsubsection{Results on CIFAR-100}
In addition, we also evaluate STTN on CIFAR-100 dataset. We compare our STTN with a strong multi-bit baseline CBCN~\cite{gu2019bonn}. CBCN replaces each convolution layer with several parallel binarized convolution layers. For fair comparisons, we use the same architecture as CBCN (ResNet-18 with 32-64-128-256 kernel stage). Note that in CBCN, the number of channels in one layer is $4\times$. Table~\ref{tab:res18_cifar100} shows our results that although CBCN uses $4\times$ channels than ours, we obtain higher accuracy with fewer computations. 
From this experiment, we argue that ternary networks can be considered before resorting to multi-bit methods.

\begin{table}
\centering
\begin{tabular}{llr}
\toprule
Model           &Kernel Stage                      &Accuracy($\%$)  \\
\midrule
Float(32+32)    &32-64-128-256                      &73.62 \\
\midrule
CBCN(1+1)       &(32-64-128-256)$\times$4            &70.07                     \\
Ours(2+2)       &32-64-128-256                    &\textbf{72.10}            \\
\bottomrule
\end{tabular}
\caption{Accuracy on CIFAR-100 with ResNet-18 (32-64-128-256). Our ternary networks can outperform multi-bit method significantly.}
\label{tab:res18_cifar100}
\end{table}

\subsubsection{Results on ImageNet}
For the large-scale dataset, we evaluate our STTN over AlexNet and ResNet-18 on ImageNet. We compare our method with several exiting state-of-the-art low-bit quantization methods: 1) only quantizing weights: BWN~\cite{courbariaux2015binaryconnect}, TWN~\cite{li2016ternary} and TTQ~\cite{zhu2016trained}; 2) quantizing both weights and activations: XNOR~\cite{rastegari2016xnor}, Bi-Real~\cite{liu2018bi}, ABC~\cite{lin2017towards}, 
TBN~\cite{wan2018tbn}, HWGQ~\cite{cai2017deep}, PACT~\cite{choi2018pact}, RTN~\cite{li2019rtn} and FATNN~\cite{chen2021fatnn}.

The overall results based on AlexNet and ResNet-18 are shown in Tabel~\ref{tab:alex_imagenet} and~\ref{tab:res18_imagenet}. We highlight our accuracy improvement (up to \textbf{15$\%$} absolute improvement compared with XNOR-Net and up to \textbf{1.7$\%$} compared with state-of-the-art ternary models, \textbf{without pre-training}). These results show that the STTN outperforms the best previous ternary methods. Such improvement indicates that our soft threshold significantly benefits extreme low-bit networks.

What's more, compared with PACT and RTN, we highlight additional improvements apart from accuracy: 1) Both PACT and RTN introduce extra floating point parameters (the activation clipping level parameter in PACT and reparameterized scale/offset in RTN) into the networks, which needs extra storage space and computation. 2) The extra introduced learnable parameters in PACT and RTN need careful manual adjustments, such as learning rate, weight decay and so on. Extensive manual tuning have to be tested for different networks on different datasets. However, our method is free of extra hyper-parameters to be tuned. Further more, we argue that our method can combine with those methods for further accuracy improvement. 3) For RTN, they argue that initialization from pre-trained full-precision models is vitally important for their methods (since the small architecture modification such as changing the order of BN and Conv layer in low-bit quantization, full-precision models released by open model zoo can not be used directly).
It is widely known that minor architecture modification (i.g. changing the order of BN and Conv) is needed in low-bit quantization. Therefore float models released by open-source model zoo can not be used. Most previous works first train full-precision weights from scratch and then use it to initialize the low-bit networks. Compared to prior works, pre-trained full precision models are not needed in our method thus much training time can be saved. And we show that by eliminating the hard threshold constraint, training from scratch can still obtain state-of-the-art results.

\begin{table}
\centering
\begin{tabular}{lccc}
\toprule
Model          &Bit-width            &Top-1($\%$)       &Top-5($\%$)\\
\midrule
XNOR           &1+1                 &44.2              &69.2    \\
TBN            &1+2                 &49.7              &74.2    \\
HWGQ           &2+2                 &52.7              &76.3    \\
PACT$^*$    &2+2                 &55.0              &$-$     \\
RTN            &2+2                 &53.9              &$-$     \\
Ours           &2+2                 &\textbf{55.6}     &\textbf{78.9}  \\
\bottomrule
\end{tabular}
\caption{Comparison with the state-of-the-art methods on ImageNet with AlexNet. "$-$" means the accuracy is not reported. "$*$" indicates the networks use quaternary values instead of ternary values for 2 bits representation.}
\label{tab:alex_imagenet}
\end{table}


\begin{table}
\centering
\begin{tabular}{lcccc}
\toprule
Model          &Bit-width       &Top-1($\%$)       &Top-5($\%$)\\
\midrule
Floating  &32+32                      &69.3              &89.2 \\
\midrule
TWN            &2+32            &61.8              &84.2    \\
TWN$^{**}$     &2+32            &65.3              &86.2    \\
TTQ$^{**}$     &2+32            &66.6              &87.2    \\
RTN$^{***}$    &2+32            &68.5              &$-$     \\
Ours           &2+32            &\textbf{68.8}      &\textbf{88.3}  \\
Ours (update)  &2+32            &\textbf{69.2}      &\textbf{89.3}  \\
\midrule
XNOR           &1+1             &51.2              &73.2    \\
Bi-Real$^{***}$ &1+1             &56.4              &79.5    \\
ABC-5          &(1$\times$5)+(1$\times$5) &65.0       &85.9    \\
TBN            &1+2             &55.6              &79.0    \\
HWGQ           &1+2             &56.1              &79.7    \\
PACT$^*$       &2+2             &64.4              &$-$     \\
RTN$^{***}$    &2+2             &64.5              &$-$     \\
FATNN          &2+2             &66.0              &86.4    \\
Ours           &2+2             &\textbf{66.2}     &\textbf{86.4} \\
Ours (update)  &2+2             &\textbf{68.2}     &\textbf{87.9} \\
\bottomrule
\end{tabular}
\caption{Comparison with the state-of-the-art methods on ImageNet with ResNet-18. "$*$" indicates the model uses quaternary values instead of ternary values for 2 bits representation. "$**$" indicates the filter number of the network is $1.5\times$. "$***$" indicates the model needs full-precision pre-trained models to initialize. "$\times$" in "ABC-5" denotes multi-bit networks with multi-branch.}
\label{tab:res18_imagenet}
\end{table}

\section{Conclusion}
In this paper, we propose a simple yet effective ternarization method, Soft Threshold Ternary Networks. We divide previous ternary works into two catalogues and show that their hard threshold is suboptimal. By simply replacing the original ternary kernel with two parallel binary kernels at training, our model can automatically determine which weights to be -1/0/1 instead of depending on a hard threshold. Experiments on various datasets show that STTN dramatically outperforms current state-of-the-arts, lowering the performance gap between full-precision networks and extreme low bit networks.

\bibliographystyle{named}
\bibliography{ijcai20}

\begin{thebibliography}{}

\bibitem[\protect\citeauthoryear{Cai \bgroup \em et al.\egroup
  }{2017}]{cai2017deep}
Zhaowei Cai, Xiaodong He, Jian Sun, and Nuno Vasconcelos.
\newblock Deep learning with low precision by half-wave gaussian quantization.
\newblock pages 5406--5414, 2017.

\bibitem[\protect\citeauthoryear{Chen \bgroup \em et al.\egroup
  }{2021}]{chen2021fatnn}
Peng Chen, Bohan Zhuang, and Chunhua Shen.
\newblock Fatnn: Fast and accurate ternary neural networks.
\newblock In {\em Proceedings of the IEEE/CVF International Conference on
  Computer Vision}, pages 5219--5228, 2021.

\bibitem[\protect\citeauthoryear{Choi}{2018}]{choi2018pact}
Jungwook Choi.
\newblock Pact: Parameterized clipping activation for quantized neural
  networks.
\newblock {\em arXiv: Computer Vision and Pattern Recognition}, 2018.

\bibitem[\protect\citeauthoryear{Courbariaux \bgroup \em et al.\egroup
  }{2014}]{courbariaux2014low}
Matthieu Courbariaux, Jean-Pierre David, and Yoshua Bengio.
\newblock Low precision storage for deep learning.
\newblock {\em arXiv preprint arXiv:1412.7024}, 2014.

\bibitem[\protect\citeauthoryear{Courbariaux \bgroup \em et al.\egroup
  }{2015}]{courbariaux2015binaryconnect}
Matthieu Courbariaux, Yoshua Bengio, and Jeanpierre David.
\newblock Binaryconnect: training deep neural networks with binary weights
  during propagations.
\newblock pages 3123--3131, 2015.

\bibitem[\protect\citeauthoryear{Courbariaux \bgroup \em et al.\egroup
  }{2016}]{courbariaux2016binarized}
Matthieu Courbariaux, Itay Hubara, Daniel Soudry, Ran El-Yaniv, and Yoshua
  Bengio.
\newblock Binarized neural networks: Training deep neural networks with weights
  and activations constrained to+ 1 or-1.
\newblock {\em arXiv preprint arXiv:1602.02830}, 2016.

\bibitem[\protect\citeauthoryear{Darabi \bgroup \em et al.\egroup
  }{2018}]{darabi2018bnn+}
Sajad Darabi, Mouloud Belbahri, Matthieu Courbariaux, and Vahid~Partovi Nia.
\newblock Bnn+: Improved binary network training.
\newblock {\em arXiv preprint arXiv:1812.11800}, 2018.

\bibitem[\protect\citeauthoryear{Gupta \bgroup \em et al.\egroup
  }{2015}]{gupta2015deep}
Suyog Gupta, Ankur Agrawal, Kailash Gopalakrishnan, and Pritish Narayanan.
\newblock Deep learning with limited numerical precision.
\newblock In {\em International Conference on Machine Learning}, pages
  1737--1746, 2015.

\bibitem[\protect\citeauthoryear{Han \bgroup \em et al.\egroup
  }{2015}]{han2015deep}
Song Han, Huizi Mao, and William~J Dally.
\newblock Deep compression: Compressing deep neural networks with pruning,
  trained quantization and huffman coding.
\newblock {\em arXiv preprint arXiv:1510.00149}, 2015.

\bibitem[\protect\citeauthoryear{He \bgroup \em et al.\egroup
  }{2019}]{he2019filter}
Yang He, Ping Liu, Ziwei Wang, Zhilan Hu, and Yi~Yang.
\newblock Filter pruning via geometric median for deep convolutional neural
  networks acceleration.
\newblock In {\em Proceedings of the IEEE Conference on Computer Vision and
  Pattern Recognition}, pages 4340--4349, 2019.

\bibitem[\protect\citeauthoryear{Hinton \bgroup \em et al.\egroup
  }{2015}]{hinton2015distilling}
Geoffrey Hinton, Oriol Vinyals, and Jeff Dean.
\newblock Distilling the knowledge in a neural network.
\newblock {\em arXiv preprint arXiv:1503.02531}, 2015.

\bibitem[\protect\citeauthoryear{Howard \bgroup \em et al.\egroup
  }{2017}]{howard2017mobilenets}
Andrew Howard, Menglong Zhu, Bo~Chen, Dmitry Kalenichenko, Weijun Wang, Tobias
  Weyand, M~Andreetto, and Hartwig Adam.
\newblock Mobilenets: Efficient convolutional neural networks for mobile vision
  applications.
\newblock {\em arXiv: Computer Vision and Pattern Recognition}, 2017.

\bibitem[\protect\citeauthoryear{Jaderberg \bgroup \em et al.\egroup
  }{2014}]{jaderberg2014speeding}
Max Jaderberg, Andrea Vedaldi, and Andrew Zisserman.
\newblock Speeding up convolutional neural networks with low rank expansions.
\newblock {\em arXiv preprint arXiv:1405.3866}, 2014.

\bibitem[\protect\citeauthoryear{kri}{}]{krizhevsky2012imagenet}


\bibitem[\protect\citeauthoryear{Li \bgroup \em et al.\egroup
  }{2016}]{li2016ternary}
Fengfu Li, Bo~Zhang, and Bin Liu.
\newblock Ternary weight networks.
\newblock {\em arXiv preprint arXiv:1605.04711}, 2016.

\bibitem[\protect\citeauthoryear{Li \bgroup \em et al.\egroup
  }{2019}]{li2019rtn}
Yuhang Li, Xin Dong, Sai~Qian Zhang, Haoli Bai, Yuanpeng Chen, and Wei Wang.
\newblock Rtn: Reparameterized ternary network.
\newblock {\em arXiv preprint arXiv:1912.02057}, 2019.

\bibitem[\protect\citeauthoryear{Lin \bgroup \em et al.\egroup
  }{2017a}]{lin2017focal}
Tsung-Yi Lin, Priya Goyal, Ross Girshick, Kaiming He, and Piotr Doll{\'a}r.
\newblock Focal loss for dense object detection.
\newblock In {\em Proceedings of the IEEE international conference on computer
  vision}, pages 2980--2988, 2017.

\bibitem[\protect\citeauthoryear{Lin \bgroup \em et al.\egroup
  }{2017b}]{lin2017towards}
Xiaofan Lin, Cong Zhao, and Wei Pan.
\newblock Towards accurate binary convolutional neural network.
\newblock In {\em Advances in Neural Information Processing Systems}, pages
  345--353, 2017.

\bibitem[\protect\citeauthoryear{Liu \bgroup \em et al.\egroup
  }{2018}]{liu2018bi}
Zechun Liu, Baoyuan Wu, Wenhan Luo, Xin Yang, Wei Liu, and Kwang-Ting Cheng.
\newblock Bi-real net: Enhancing the performance of 1-bit cnns with improved
  representational capability and advanced training algorithm.
\newblock In {\em Proceedings of the European Conference on Computer Vision
  (ECCV)}, pages 722--737, 2018.

\bibitem[\protect\citeauthoryear{Liu \bgroup \em et al.\egroup
  }{2019}]{gu2019bonn}
ChunLei Liu, Wenrui Ding, Xin Xia, Baochang Zhang, Jiaxin Gu, Jianzhuang Liu,
  Rongrong Ji, and Doermann David.
\newblock Circulant binary convolutional networks: Enhancing the performance of
  1-bit dcnns with circulant back propagation.
\newblock In {\em IEEE Conference on Computer Vision and Pattern Recognition},
  2019.

\bibitem[\protect\citeauthoryear{Rastegari \bgroup \em et al.\egroup
  }{2016}]{rastegari2016xnor}
Mohammad Rastegari, Vicente Ordonez, Joseph Redmon, and Ali Farhadi.
\newblock Xnor-net: Imagenet classification using binary convolutional neural
  networks.
\newblock In {\em European Conference on Computer Vision}, pages 525--542.
  Springer, 2016.

\bibitem[\protect\citeauthoryear{Tang \bgroup \em et al.\egroup
  }{2017}]{tang2017train}
Wei Tang, Gang Hua, and Liang Wang.
\newblock How to train a compact binary neural network with high accuracy?
\newblock In {\em Thirty-First AAAI Conference on Artificial Intelligence},
  2017.

\bibitem[\protect\citeauthoryear{Wan \bgroup \em et al.\egroup
  }{2018}]{wan2018tbn}
Diwen Wan, Fumin Shen, Li~Liu, Fan Zhu, Jie Qin, Ling Shao, and Heng Tao~Shen.
\newblock Tbn: Convolutional neural network with ternary inputs and binary
  weights.
\newblock In {\em Proceedings of the European Conference on Computer Vision
  (ECCV)}, pages 315--332, 2018.

\bibitem[\protect\citeauthoryear{Wang \bgroup \em et al.\egroup
  }{2018}]{wang2018two}
Peisong Wang, Qinghao Hu, Yifan Zhang, Chunjie Zhang, Yang Liu, and Jian Cheng.
\newblock Two-step quantization for low-bit neural networks.
\newblock In {\em Proceedings of the IEEE Conference on Computer Vision and
  Pattern Recognition}, pages 4376--4384, 2018.

\bibitem[\protect\citeauthoryear{Zhou \bgroup \em et al.\egroup
  }{2016}]{zhou2016dorefa}
Shuchang Zhou, Yuxin Wu, Zekun Ni, Xinyu Zhou, He~Wen, and Yuheng Zou.
\newblock Dorefa-net: Training low bitwidth convolutional neural networks with
  low bitwidth gradients.
\newblock {\em arXiv preprint arXiv:1606.06160}, 2016.

\bibitem[\protect\citeauthoryear{Zhu \bgroup \em et al.\egroup
  }{2016}]{zhu2016trained}
Chenzhuo Zhu, Song Han, Huizi Mao, and William~J Dally.
\newblock Trained ternary quantization.
\newblock {\em arXiv preprint arXiv:1612.01064}, 2016.

\bibitem[\protect\citeauthoryear{Zhu \bgroup \em et al.\egroup
  }{2019}]{zhu2019binary}
Shilin Zhu, Xin Dong, and Hao Su.
\newblock Binary ensemble neural network: More bits per network or more
  networks per bit?
\newblock In {\em Proceedings of the IEEE Conference on Computer Vision and
  Pattern Recognition}, pages 4923--4932, 2019.

\bibitem[\protect\citeauthoryear{Zoph and Le}{2016}]{zoph2016neural}
Barret Zoph and Quoc~V Le.
\newblock Neural architecture search with reinforcement learning.
\newblock {\em arXiv preprint arXiv:1611.01578}, 2016.

\end{thebibliography}
\end{document}